\documentclass[utf8]{frontiersSCNS} 

\usepackage{url,lineno,microtype,subcaption}
\usepackage[onehalfspacing]{setspace}
\usepackage{comment}
\usepackage{algorithm}
\usepackage{algorithmic}
\usepackage{placeins}

\def\keyFont{\fontsize{8}{11}\helveticabold }
\def\firstAuthorLast{Natarajan {et~al.}} 
\def\Authors{Sabhari Natarajan\,$^{1,\dagger}$ Galen Brown\,$^{2,\dagger,*}$, and Berk Calli\,$^{1,2}$}
\def\Address{$^{1}$ Manipulation and Environmental Robotics Laboratory (MER Lab), Robotics Engineering Department, Worcester Polytechnic Institute, Worcester, MA, United States \\
$^{2}$ Manipulation and Environmental Robotics Laboratory (MER Lab), Computer Science Department, Worcester Polytechnic Institute, Worcester, MA, United States \\
$\dagger$ Marked authors contributed equally to this publication and share first authorship.}

\def\corrAuthor{Galen Brown, 100 Institute Road, Mailbox G103, Worcester, MA, 01609}

\def\corrEmail{gibrown@wpi.edu}

\begin{document}
\onecolumn
\firstpage{1}

\title[Grasp Synthesis Using Active Vision]{Grasp Synthesis for Novel Objects Using Heuristic-based and Data-driven Active Vision Methods}

\author[\firstAuthorLast ]{\Authors} 
\address{} 
\correspondance{} 

\extraAuth{}

\maketitle

\begin{abstract}

\section{}
In this work, we present several heuristic-based and data-driven active vision strategies for viewpoint optimization of an arm-mounted depth camera for the purpose of aiding robotic grasping. These strategies aim to efficiently collect data to boost the performance of an underlying grasp synthesis algorithm. We created an open-source benchmarking platform in simulation (https://github.com/galenbr/2021ActiveVision), and provide an extensive study for assessing the performance of the proposed methods as well as comparing them against various baseline strategies. We also provide an experimental study with a real-world setup by utilizing an existing grasping planning benchmark in the literature. With these analyses, we were able to quantitatively demonstrate the versatility of heuristic methods that prioritize certain types of exploration, and qualitatively show their robustness to both novel objects and the transition from simulation to the real world. We identified scenarios in which our methods did not perform well and scenarios which are objectively difficult, and present a discussion on which avenues for future research show promise. 

\tiny
 \keyFont{ \section{Keywords:} active vision, self-supervised learning, reinforcement learning, grasp synthesis, benchmarking} 
\end{abstract}

\section{Introduction}


\begin{figure}[h]
    \centering
    \captionsetup{justification=centering}
    \includegraphics[width=\linewidth]{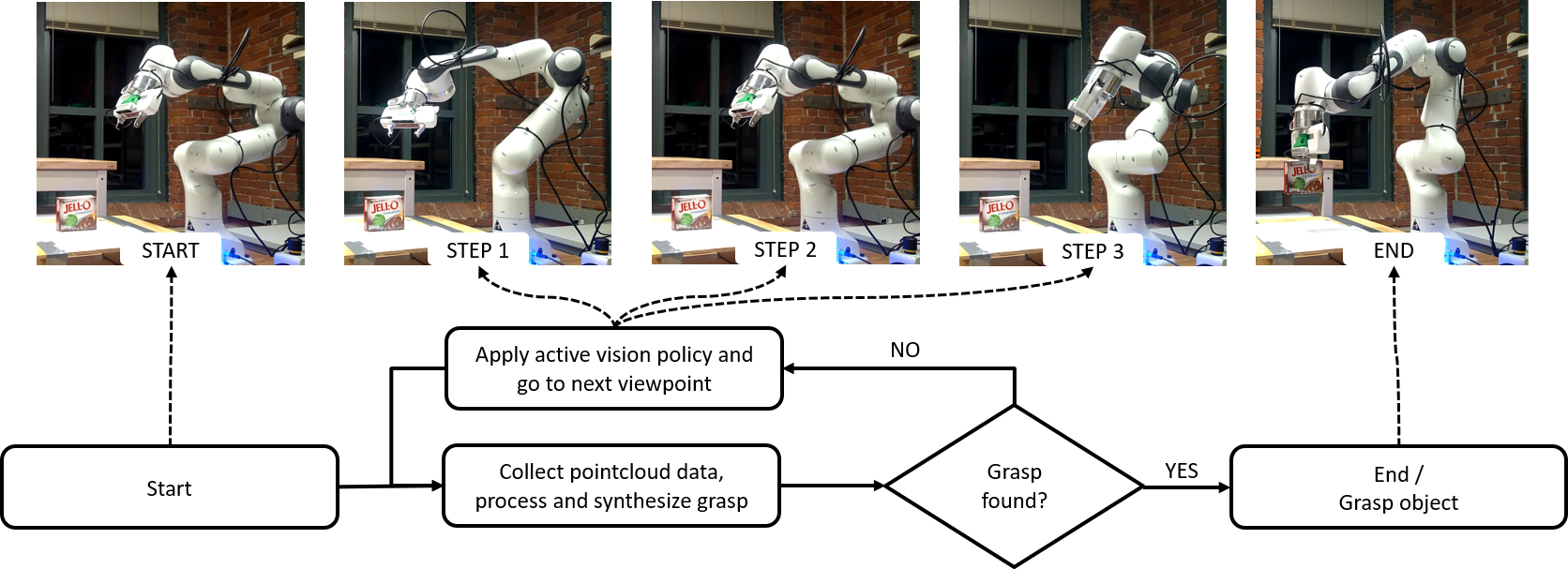}
    \caption{The 3D Heuristic policy guiding the camera and finding the grasp for a object}
    \label{fig_intro}
\end{figure}

Robotic grasping is a vital capability for many tasks, particularly in service robotics. Most grasping algorithms use data from a single viewpoint to synthesize a grasp \citep{Caldera2018}. This approach attempts to create a single, master algorithm that is useful for all objects in all situations. Nevertheless, these algorithms tend to suffer when the viewpoint of the vision sensor is different than the images used in training \citep{Viereck2017}. Additionally, many graspable objects have observation angles that are ``singular" from which no grasp can be synthesized: For example, if an object has only one graspable surface, which is self-occluded from the current viewpoint of the camera, the grasp synthesis algorithm would either fail to find any grasps or would need to rely on assumptions that might not always hold, and therefore lead to an unsuccessful grasp attempt. 

The issues of the single viewpoint approaches can be addressed via active vision frameworks, i.e. by actively moving the camera and collecting more data about the task. At one end of this spectrum is collecting data to obtain a complete 3D model of the object. This approach is slow, difficult to carry out in the real world, and vulnerable to misalignment if conditions change during or after data collection \citep{Lakshminarayanan2017}. Our aim is to develop active vision strategies that can efficiently collect data with brief motions and allow the grasp synthesis algorithms to find sufficiently good grasps as quickly as possible. It is shown in the grasping literature that even algorithms tailored for single viewpoints can have substantial performance boost even with very simple data collection procedures \citep{Viereck2017}. Utilizing active vision for robotic grasping has several avenues for optimization: the exploration algorithm, the data analysis, and the grasping algorithm are all open questions. 

In this work, we present a wide variety of exploration algorithms along with an extensive simulation and real-world experiment analysis. Figure \ref{fig_intro} shows how an active vision policy explores different objects. In simulation, we created benchmarks to assess not only whether our policies do better than random but to measure how close each approach comes to optimal behavior for each object. In the real-world experiments, we have adopted an existing grasp planning benchmark \citep{Bekiroglu2020}, and assess how well the simulation performances translate to real systems.

Our exploration algorithms can be split into heuristic and machine learning approaches. In our heuristics, we attempt to identify simple properties of the visual data that are reliable indicators of effective exploration directions. These approaches use estimates of how many potentially occluded grasps lie in each direction. For machine learning, we used self-supervised and Q-learning based approaches. We compare the performance of these methods against three baseline algorithms. The baselines are random motion (as the worst case algorithm), naive straight forward motion (as a simple algorithm more complex efforts should outperform), and breadth-first-search (as the absolute ceiling on possible performance). The last is particularly important: because in simulation we could exhaustively test each possible exploration path, we can say with certainty what the shortest possible exploration path that leads to a working grasp is. We also present a comparison study to another active vision-based algorithm, i.e. \citep{Arruda2016}, which provides, to the best of our knowledge, the closest strategy to ours in the literature.

To summarize, the contribution of our work is as follows:
\begin{enumerate}
    \item We present two novel heuristic-based viewpoint optimization methods.
    \item We provide a novel Q-learning based approach for achieving an exploration policy.
    \item We provide an open-source simulation platform (https://github.com/galenbr/2021ActiveVision) to develop new active vision algorithms and benchmark them.
    \item We present an extensive simulation and experimental analysis, assessing and comparing the performance of 5 active vision methods against 3 baseline strategies.
\end{enumerate}
Taken together, these allow us to draw new conclusions not only about how well our algorithms work now, but how much it would be possible to improve them. 

\section{Related Works}
Adapting robotic manipulation algorithms to work in an imperfect and uncertain world is a central concern of the robotics field, and an overview of modern approaches is given by \cite{Wang2020}. For the use of active vision to address this problem, there has been research into both algorithmic \citep{Calli2011,Arruda2016} and data-driven methods \citep{Paletta2000, Viereck2017,Calli2018,Rasolzadeh2010}, with more recent works tending to favor data-driven approaches \citep{Caldera2018}. In particular, the work in \citep{Viereck2017} demonstrated that active vision algorithms have the potential to outperform state of the art single-shot grasping algorithms. 

\cite{Calli2011} proposed an algorithmic active vision strategy for robotic grasping, extending 2D grasp stability metrics to 3D space. As an extension of that work \citep{Calli2018}, the authors utilized local optimizers for systematic viewpoint optimization using 2D images. \cite{Arruda2016} employs a probabilistic algorithm whose core approach is the most similar to our heuristics presented in Section~\ref{ssec:heuristic-policies}. Our approaches differ in focus, with \cite{Arruda2016} selects viewpoints based on estimated information gain as a proxy for finding successful grasps, while we prioritize grasp success likelihood and minimizing distance traveled. In our simulation study, we implemented a version of their algorithm and included it our comparison analysis.

The data-driven approach presented in \cite{Viereck2017} avoided the problem of labeled data by automating data labeling using state of the art single shot grasp synthesis algorithms. They then used machine learning to estimate the direction of the nearest grasp along a view-sphere, and performed gradient descent along the vector field of grasp directions. This has the advantage of being continuous and fast, but did not fit in our discrete testing framework \citep{Viereck2017}. All data-driven methods analysed in this paper utilize a similar self-supervised learning framework due to its significant easiness in training. 

One of our data-driven active vision algorithms utilize the reinforcement learning framework. A similar strategy for active vision is used by \cite{Paletta2000} to estimate an information gain maximizing strategy for object recognition. We not only extend Q-learning to grasping, but do away with the intermediary information gain heuristic in reinforcement learning. Instead we penalize our reinforcement approach for each step it takes that does not find a grasp, incentivizing short, efficient paths.

Two of the data-driven methods in this paper is based on the general strategy in our prior work in \cite{Calli2018}. In that work, we presented a preliminary study was presented in simulation. In this paper, we present one additional variant of this strategy, and present a more extended simulation analysis.

\cite{Gallos2019}, while focused on classification rather than grasping, heavily influenced our theoretical concerns and experimental design. Their paper argues that contemporary machine learning based active vision techniques outperform random searches but that this is too low a bar to call them useful, and demonstrate that none of the methods they implemented could outperform the simple heuristic of choosing a direction and moving along it in large steps. Virtually all active vision literature (e.g. \cite{DeCroon2009,Ammirato2017}) compares active vision approaches to random approaches or single shot state of the art algorithms. While there has been research on optimality comparison in machine vision \citep{Karasev}, to the best of our knowledge it has never been extended to 3D active vision, much less active vision for grasp synthesis. Our simulation benchmarks are an attempt to not only extend their approach to grasping, but to quantify how much improvement over the best performing algorithms remains possible.



\section{Overview}

\begin{figure}[h]
    \centering
    \captionsetup{justification=centering}
    \includegraphics[width=0.8\linewidth]{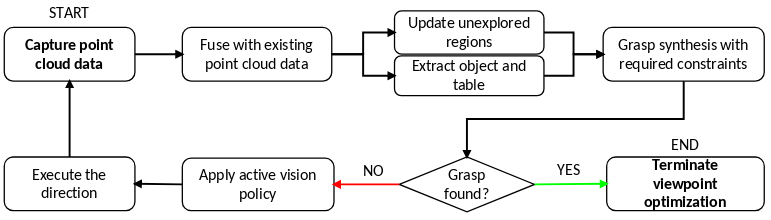}
    \caption{The active vision based grasp synthesis pipeline}
    \label{fig_methodology}
\end{figure}

The proposed active vision based grasp synthesis pipeline is represented in Figure \ref{fig_methodology}. It starts with collecting environment information from a viewpoint and fusing with the previously known information about the environment (except for the first viewpoint captured). The object and table data are extracted, apart from updating the regions which have not been explored (unexplored regions) by the camera yet. This processed data is used in the grasp synthesis and active vision policies which will be explained in the further parts of the paper. An attempt is made to synthesize a grasp with the available data, and if it fails, the active vision policy is called to guide the camera to its next viewpoint after which the process repeats until the grasp has been found.

\FloatBarrier

\subsection{Workspace description}

\begin{figure}[h]
    \centering
    \captionsetup{justification=centering}
    \includegraphics[width=0.3\linewidth]{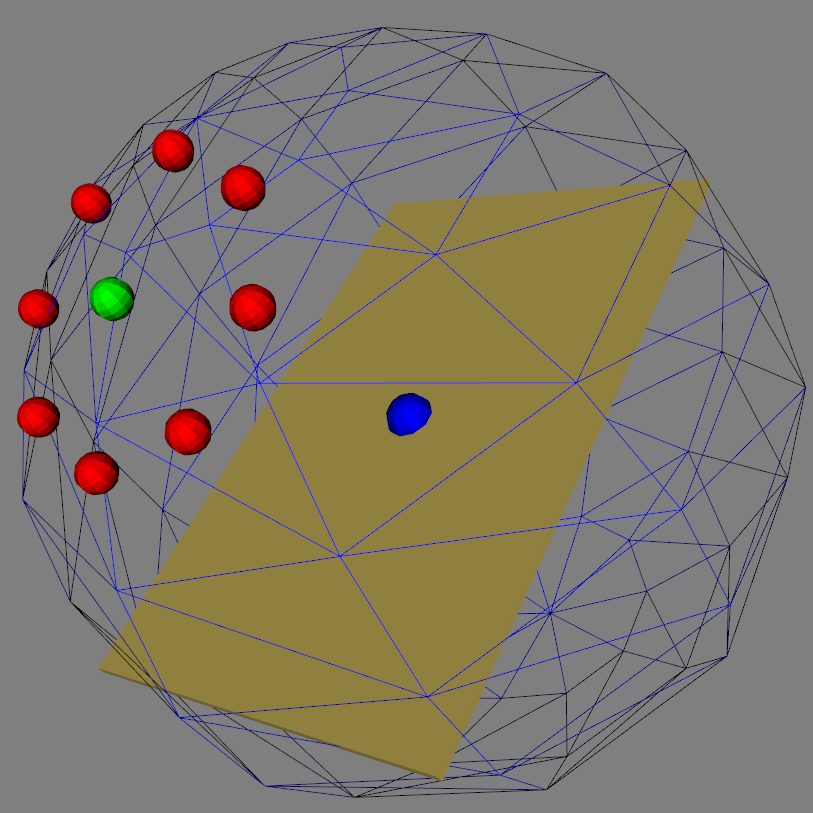}
    \caption{Viewsphere and its next steps with parameters $v_{r}$ = 0.4m,  $v_{s}$ = 20\textdegree and  $v_{d}$ = 8. The blue sphere is the expected position of the object, green sphere the current camera position and red one the next steps it can take}
    \label{fig_viewsphere}
\end{figure}

We assume an eye-in-hand system that allows us to move the camera to any viewpoint within the manipulator workspace. To reduce the dimension of active vision algorithm's action space, the camera movement is constrained to move along a viewsphere, always pointing towards and centered around the target object (a common strategy also adopted in \cite{Paletta2000,Arruda2016,Calli2018a}). The radius of the viewsphere ($v_{r}$) is set based on the manipulator workspace and sensor properties. In the viewsphere, movements are discretized into individual steps with two parameters, step-size ($v_{s}$) and number of directions ($v_{d}$). Figure \ref{fig_viewsphere} shows the workspace we use with $v_{r}$ = 0.4m,  $v_{s}$ = 20\textdegree and  $v_{d}$ = 8 (N,NE,E,SE,S,SW,W,NW). In our implementation, we use a Intel Realsense D435i as the camera on Franka Emika Panda arm for our eye-in-hand system. 

\subsection{Point Cloud Processing and Environment modelling}
\begin{figure}[h]
    \centering
    \captionsetup{justification=centering}
    \includegraphics[width=0.9\linewidth]{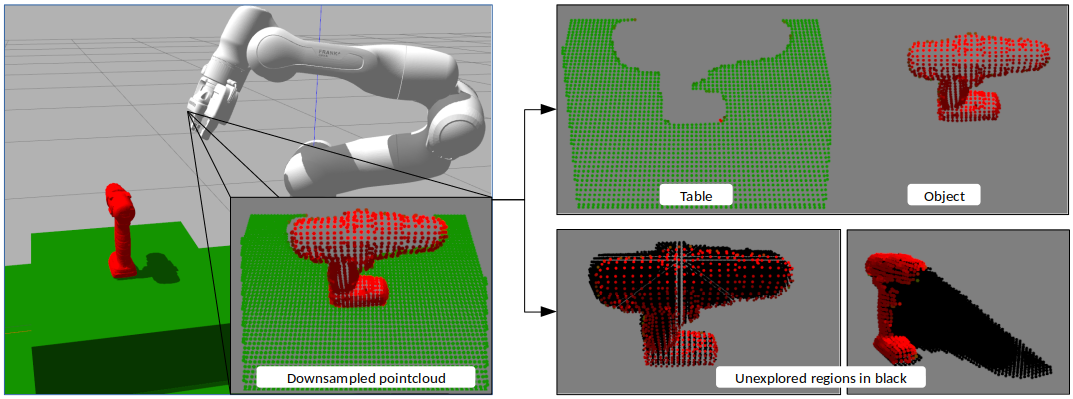}
    \caption{Example with power drill as object showing the processed pointclouds. Left : Environment as seen by the camera, right-top : Extracted object and table, right bottom : The unexplored regions of the environment}
    \label{fig_obj_modelling}
\end{figure}

The point cloud data received from the camera is downsampled before further processing to reduce sensor noise and to speed up the execution time. Figure \ref{fig_obj_modelling} shows the environment as seen by the camera after downsampling.

Sample Consensus based plane segmentation techniques in Point Cloud Library \citep{Rusu_ICRA2011_PCL} is used to extract the table information from the scene following which the points above the table are extracted to be marked as object points. As mentioned previously, identifying the unexplored regions is required for grasp synthesis as well as the active vision policies. For this purpose, the region surrounding the object is populated with an evenly spaced point cloud and then sequentially checked the determine which points are occluded. While a common visibility check approach is ray-tracing, it is a computationally intensive and time consuming process. Instead, we take advantage of the organised nature of the point cloud data, and use the camera intrinsic matrix ($K$) to project the 3D points ($X$) to the image plane (Eqn. \ref{eqn_projection}), and compare the depth values of X and the point present in the environment at pixel coordinate $X_{p}$. This approach leads to a much faster computation. The two images on the bottom right of Figure \ref{fig_obj_modelling} show the unexplored region generated for the the drill object.

\begin{equation}
    \label{eqn_projection}
    Projected\;pixels\;:\;X_{p} = K X / z_{0},\;
    where\;
        K = \begin{pmatrix}  
            f_{x} & 0 & pp_{x} \\
            0 & f_{y} & pp_{y} \\
            0 & 0 & 1
            \end{pmatrix}\;and\;
        X=\begin{pmatrix} x_{0} \\ y_{0} \\ z_{0} \end{pmatrix}
\end{equation}

With every new viewpoint the camera is moved to, the newly acquired point cloud is fused with the existing environment data and the process is repeated to extract the object data and update the unexplored regions.

\subsection{Grasp synthesis}
Synthesising a successful grasp is an important part in this pipeline. Essentially, any grasp synthesis algorithm can be used in this methodology. However, these algorithms are naturally preferred to be fast (since they would be run multiple times per grasp), and be able to work with stitched point clouds. Most data-driven approaches in the literature are trained with single-view point clouds, and might not designed to perform well with stitched object data. Instead, we use a force-closure-based approach similar to \citep{Calli2018a}, but with following two additional constraints to make the grasps more reliable:
\begin{enumerate}
    \item Contact patch constraint: Based on the known gripper contact area and the point surrounding the point under consideration, the contact patch area is calculated by projecting the points to the contact plane. This area should be higher than a threshold for both points in the candidate. 
    \item Curvature constraint : The curvature of both the points should be less than a defined threshold.
\end{enumerate}
On the stitched object data, we search for point pairs that satisfy our criteria: The angle between the normal vectors of the two grasp contact points is the grasp quality metric used. With both vectors pointing directly towards each other we will have the highest quality of 180, with the lowest possible value being 0. A minimum threshold of 150 is set in this study. The unexplored region point cloud is used at this stage to do the collision check before selecting the best available grasp. The grasps close to the line of gravity and high grasp quality are given higher preference during the grasp selection process. Any grasps that intersect with unexplored regions are omitted and therefore the grasp candidates do not make assumptions on the object shape (since they use only the already seen surfaces).

Next we explain the active vision policies designed and utilized in this paper.
\section{Active Vision Policies}

The focus of this paper is the active vision policies, which guide the eye-in-hand system to its next viewpoints. The nature of the pipeline allows us to plug in any policy which takes point clouds as its input and returns the direction to move for the next viewpoint. The policies developed and tested in this paper have been classified into three categories as follows:
\begin{enumerate}
    \item Baseline policies
    \item Heuristic policies
    \item Machine learning policies
\end{enumerate}
Each of these sets of policies are explained below.

\subsection{Baseline Policies}
As the name suggests these are a set of policies defined to serve as a baseline to compare the heuristic and machine learning policies with. The three baselines used are shown below.
    
    \subsubsection{Random Policy}
    Ignoring camera data, a random direction was selected for each step. No constraints were placed on the direction chosen, leaving the algorithm free to (for instance) oscillate infinitely between the start pose and positions one step away. This represents the worst case for an algorithm not deliberately designed to perform poorly, and all methods should be expected to perform better than it in the aggregate. This is the standard baseline in the active vision literature.

    \subsubsection{Brick Policy}
    Named after throwing a brick on the gas pedal of a car, a consistent direction (North East) was selected at each timestep. This direction was selected because early testing strongly favored it, but we make no claims that it is ideal. This policy represents the baseline algorithm which is naively designed and which any serious algorithm should be expected to outperform, but which is nonetheless effective. Any algorithm that performed more poorly than it would need well-justified situational advantages to be usable. 
    
    \subsubsection{Breadth-First-Search (BFS) Policy}
    From the starting position, an exhaustive Breadth-First-Search is performed, and an optimal path is selected. This policy represents optimal performance, as it is mathematically impossible for a discrete algorithm to produce a shorter path from the same start point. No discrete method can exceed its performance, but measuring how close each method comes to it gives us an objective measure of each method’s quality in each situation. 

With baselines defined, we will now discuss the other categories starting with heuristics.
    
\subsection{Heuristic Policies} \label{ssec:heuristic-policies}
The idea behind the heuristic policy is to choose the best possible direction after considering next available viewpoints. The metric used to define the quality of each of the next viewpoints is a value proportional to the unexplored region visible from a given viewpoint.
    
    \subsubsection{2D Heuristic Policy}
    The viewpoint quality is calculated by transforming the point clouds to the next possible viewpoints, and projecting the object and unexplored point clouds from those viewpoints onto a image plane using the camera’s projection matrix. This process has the effect of making the most optimistic estimation for exploring unexplored regions; it assumes no new object points will be discovered from the new viewpoint. Since the point clouds were downsampled, their projected images were dilated to generate closed surfaces. The 2D projections are then overlapped to calculate the size of the area not occluded by the object. The direction for which the most area of unexplored region is revealed is then selected. Figure \ref{fig_2D_3D_Heuristic} shows a illustration with the dilated projected surfaces and the calculated non-occluded region. The 2D Heuristic policy is outlined in Algorithm \ref{alg:2DHeuristic}.
    
    \begin{algorithm}
    \caption{2D Heuristic policy}
    \label{alg:2DHeuristic}
    \begin{algorithmic} 
    \REQUIRE $obj \leftarrow$ Object point cloud
    \REQUIRE $unexp \leftarrow$ Unexplored point cloud
    \FORALL{$viewpoint \in$ next possible viewpoints}
        \IF{viewpoint within manipulator workspace}
            \STATE $obj\_trf \leftarrow$ Transform $obj$ to viewpoint
            \STATE $obj\_proj \leftarrow$ Project $obj\_trf$ onto image plane (B/W image) and dilate
            
            \STATE $unexp\_trf \leftarrow$ Transform $unexp$ to viewpoint
            \STATE $unexp\_proj \leftarrow$ Project $unexp\_trf$ onto image plane (B/W image) and dilate
            
            \STATE $non\_occ\_unexp\_proj \leftarrow unexp\_proj - obj\_proj$
        \ENDIF
        \STATE Record the number of white pixels in $non\_occ\_unexp\_proj$
    \ENDFOR
    \STATE Choose the direction with maximum white pixels
    \end{algorithmic}
    \end{algorithm}
    
    While this heuristic is computational efficient, it considers the 2D projected area, leading it to, at times, prefer wafer thin slivers with high projected area over deep blocks with low projected area. Additionally, it is agnostic to the grasping goal, and only focuses on maximizing the exploration of unseen regions. 
    
    \subsubsection{3D Heuristic Policy}\label{ssec:3d-heuristic-policy}
    
    In the 3D heuristic, we focused only on the unexplored region which could lead to a potential grasp. This was done using the normal vectors of the currently visible object. Since our grasp algorithm relies on antipodal grasps, only points along the surface normals can produce grasps. We found the unexplored points within the grasp width of gripper and epsilon of those normal vectors, and discarded all other points from the unexplored point cloud.
    
\begin{figure}[h]
        \centering
        \captionsetup{justification=centering}
        \includegraphics[width=0.9\linewidth]{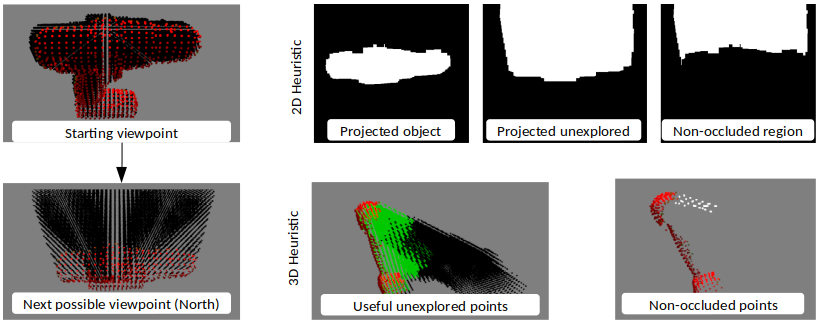}
        \caption{Set of images illustrating how the 2D and 3D Heuristics evaluate a proposed next step North with the drill object. The 3D Heuristic images have been shown from a different viewpoint for representation purposes.}
        \label{fig_2D_3D_Heuristic}
    \end{figure}

    Next, like in the 2D heuristic, we transformed the points to the next possible viewpoints. This time, instead of projecting, we used local surface reconstruction and ray-tracing to determine all the unexplored points which will not be occluded from a given viewpoint. The direction which leads to the highest number of non-occluded unexplored points is selected. This prioritizes exploring the greatest possible region of unexplored space that, based on known information, could potentially contain a grasp. If all the viewpoints after one step have very few non-occluded points the policy looks one step ahead in the same direction for each before making the decision. Figure \ref{fig_2D_3D_Heuristic} shows a illustration with the non-occluded useful unexplored region. The green points are the region of unexplored region which is considered useful based on gripper configuration. The 3DHeuristic policy is outlined in Algorithm \ref{alg:3DHeuristic}.
    
    \begin{algorithm}
    \caption{3D Heuristic policy}
    \label{alg:3DHeuristic}
    \begin{algorithmic} 
    \REQUIRE $obj \leftarrow$ Object point cloud
    \REQUIRE $unexp \leftarrow$ Unexplored point cloud
    \REQUIRE $points\_threshold \leftarrow$ Minimum number of non-occluded unexplored points needed for a new viewpoint to be considered useful
    \STATE $useful\_unexp\_trf \leftarrow$ Unexplored points with potential for a successful grasp
    \FORALL{$viewpoint \in$ next possible viewpoints}
        \IF{viewpoint within manipulator workspace}
            \STATE $obj\_trf \leftarrow$ Transform $obj$ to viewpoint
            \STATE $useful\_unexp\_trf \leftarrow$ Transform $useful_unexp$ to viewpoint
            
            \STATE $non\_occ\_useful\_unexp \leftarrow$ Check occlusion for each $useful\_unexp\_trf$ using local surface reconstruction and ray-tracing.
        \ENDIF
        \STATE Record the number of points in $non\_occ\_useful\_unexp$
    \ENDFOR
    \STATE $max\_points \leftarrow$ Maximum points seen across the possible viewpoints 
    \IF{$max\_points \leq points\_threshold$}
        \STATE Run the previous for loop with twice the step-size
    \ENDIF
    \STATE $max\_points \leftarrow$ Maximum points seen across the possible viewpoints 
    \STATE Choose the direction which has $max\_points$
    \end{algorithmic}
    \end{algorithm}
    
    \FloatBarrier

    \subsubsection{Information Gain Heuristic Policy}
    The closest approach to the heuristics presented in this paper is provided by \cite{Arruda2016}. For comparison purposes, we implemented an approximate version of their exploration policy to test our assumptions and compare it with our 3D Heuristic approach. First we defined a set of 34 viewpoints spread across the viewsphere to replicate their search space. To calculate the information gain for each viewpoint, we modified the 3D Heuristic to consider all unexplored regions as opposed to focusing on the regions with a potential grasp. Similarly the modified 3D Heuristic policy, instead of comparing the next $v_{d}$ viewpoints, compared all 34 viewpoints and used the one with the highest information gain. A simulation study was performed to compare the camera travel distance and computation times of this algorithm to our other heuristics.
    
\subsection{Machine Learning Policies}
    
Our data-driven policies utilize a fixed size state vector as input. A portion of this vector is obtained by modelling the object point cloud and unexplored regions point cloud with Height accumulated features (HAF), which was also used in \cite{Calli2018a}. We experimented with grid sizes of 5 and 7 height maps, both of which provide similar performance in our implementation, and we chose to use 5. The state vector of a given view is composed of the flattened height maps of the extracted object and the unexplored point cloud and the polar and azimuthal angle of camera in viewsphere. The size of the state vector is $2n^2+2$, where $n$ is the grid size.

    \subsubsection{Self-supervised Learning Policy}
    Following the synthetic data generation used in \citep{Calli2018a}, we generated training data by randomly exploring up to five steps in each direction three times, and choosing the shortest working path in simulation. This was repeated for 1,000 random initial poses each for two simple rectangular prisms in Figure \ref{fig_sim_train_objects}. We then applied PCA to each vector to further compress it to 26 components. We have two variations for using this data: In one variation we trained a simple logistic regression classifier to take a compressed state vector and predict the next direction to take from it. In the second variation, we trained an LDA classifier to predict the next direction from the compressed state vector. All the components used in this policy were implemented in the scikit-learn library\citep{scikit-learn}. 
    
    \begin{figure}[h]
        \centering
        \captionsetup{justification=centering}
        \includegraphics[width=0.75\linewidth]{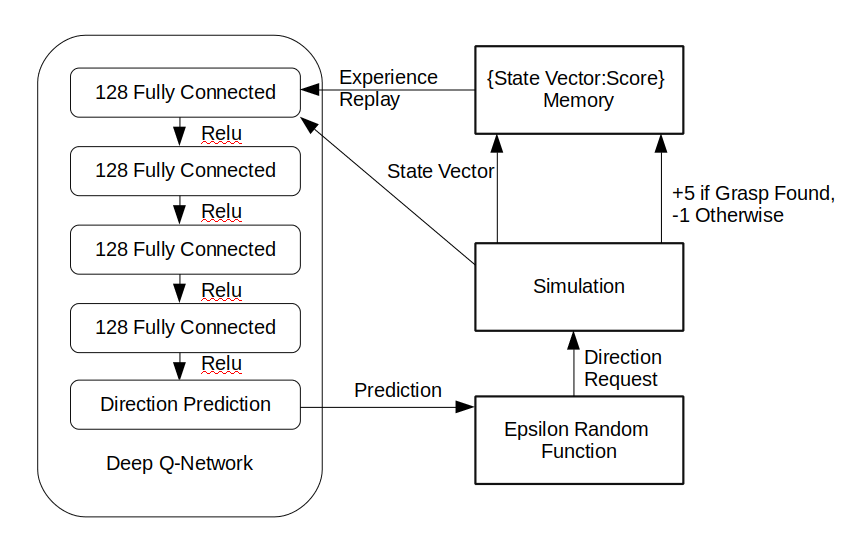}
        \caption{The Deep Q-Learning policy}
        \label{q_learning_arch}
    \end{figure}

    \subsubsection{Deep Q-Learning Policy}
\begin{figure}[h]
        \centering
        \captionsetup{justification=centering}
        \includegraphics[width=0.75\linewidth]{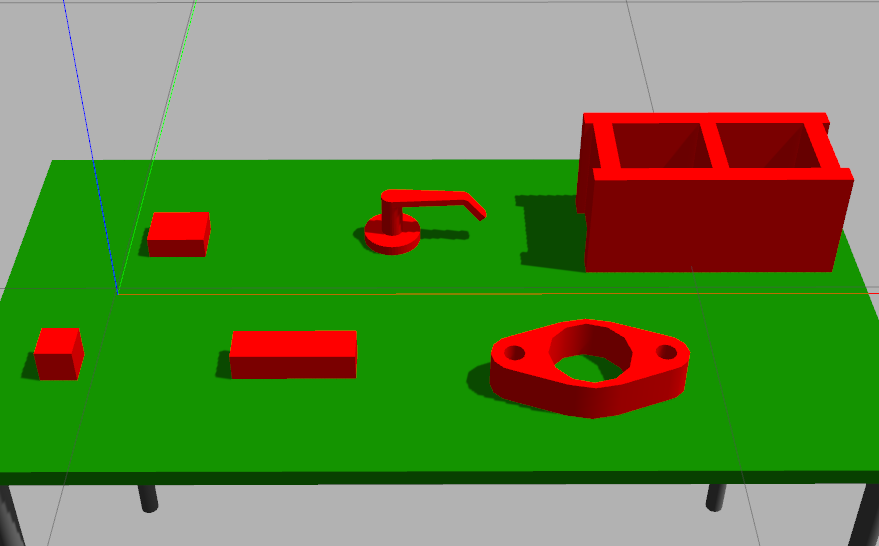}
        \caption{The set of object used for simulation training. Filenames left to right: prism 6x6x6, prism 10x8x4, prism 20x6x5, handle, gasket, cinder block. Only 
        prism 10x8x4, prism 20x6x5 were used to train the supervised learning algorithms.}
        \label{fig_sim_train_objects}
    \end{figure}

    A deep Q-Learning policy was trained to predict, for a given state vector, the next step that would lead to the shortest path to a viable grasp using Keras library tools \citep{chollet2015keras}. Four fully connected 128 dense layers and one 8 dense layer, connected by Relu transitions, formed the deep network that made the predictions. In training, an epsilon-random gate replaced the network's prediction with a random direction if a random value exceeded an epsilon value that decreased with training. The movement this function requested was then performed in simulation, and the resulting state vector and a binary grasp found metric were recorded. Once enough states had been captured, experience replay randomly selected from the record to train the Q-Network on a full batch of states each iteration. The Q-Learning was trained in simulation to convergence on all of the objects in Figure \ref{fig_sim_train_objects}, taking roughly 1,300 simulated episodes to reach convergence. We hoped that, given the relatively constrained state space and strong similarities between states, meaningful generalizations could be drawn from the training set to completely novel objects. 

For all machine learning approaches, the objects used for training were never used in testing.
\FloatBarrier
\section{Simulation and Experimental Results}
\begin{figure}[h]
    \centering
    \captionsetup{justification=centering}
    \includegraphics[width=0.9\linewidth]{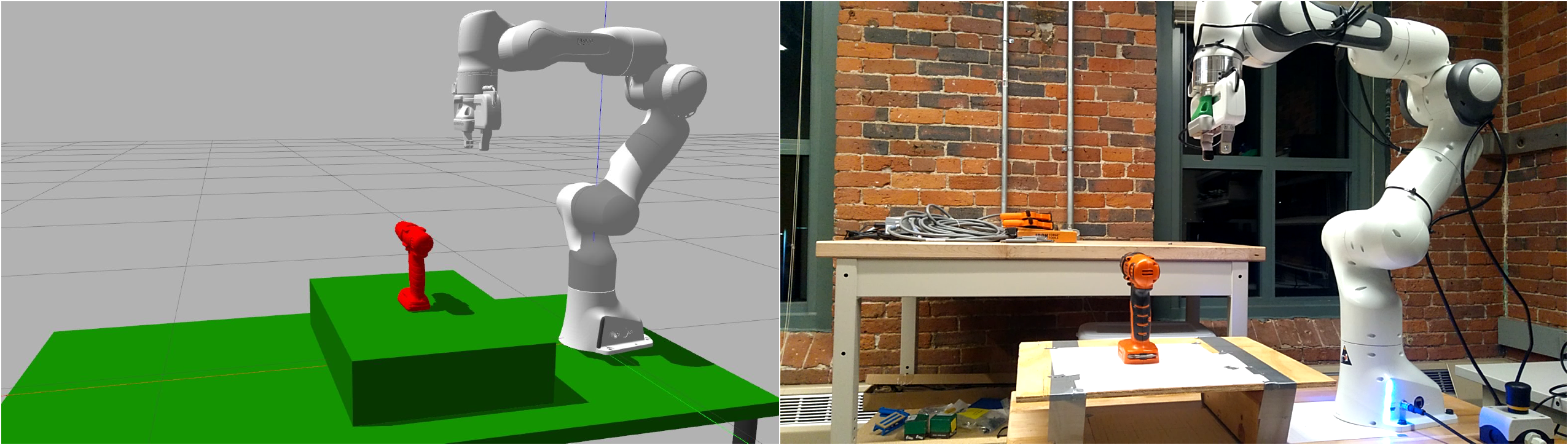}
    \caption{The setup as seen in simulation environment (left) and lab environment (right) with the YCB object power drill (ID : 35) in place}
    \label{fig_lab_sim_setup}
\end{figure}

The methodology discussed in the above section was implemented and tested in both simulation and in the real world. The setups used for the testing are shown in Figure \ref{fig_lab_sim_setup}. Maximum number of steps allowed before a experiment is restarted was set to 6 on the basis of preliminary experiments with the BFS policy.

\FloatBarrier
\subsection{Simulation Study}

\begin{figure}[h]
    \centering
    \captionsetup{justification=centering}
    \includegraphics[width=0.75\linewidth]{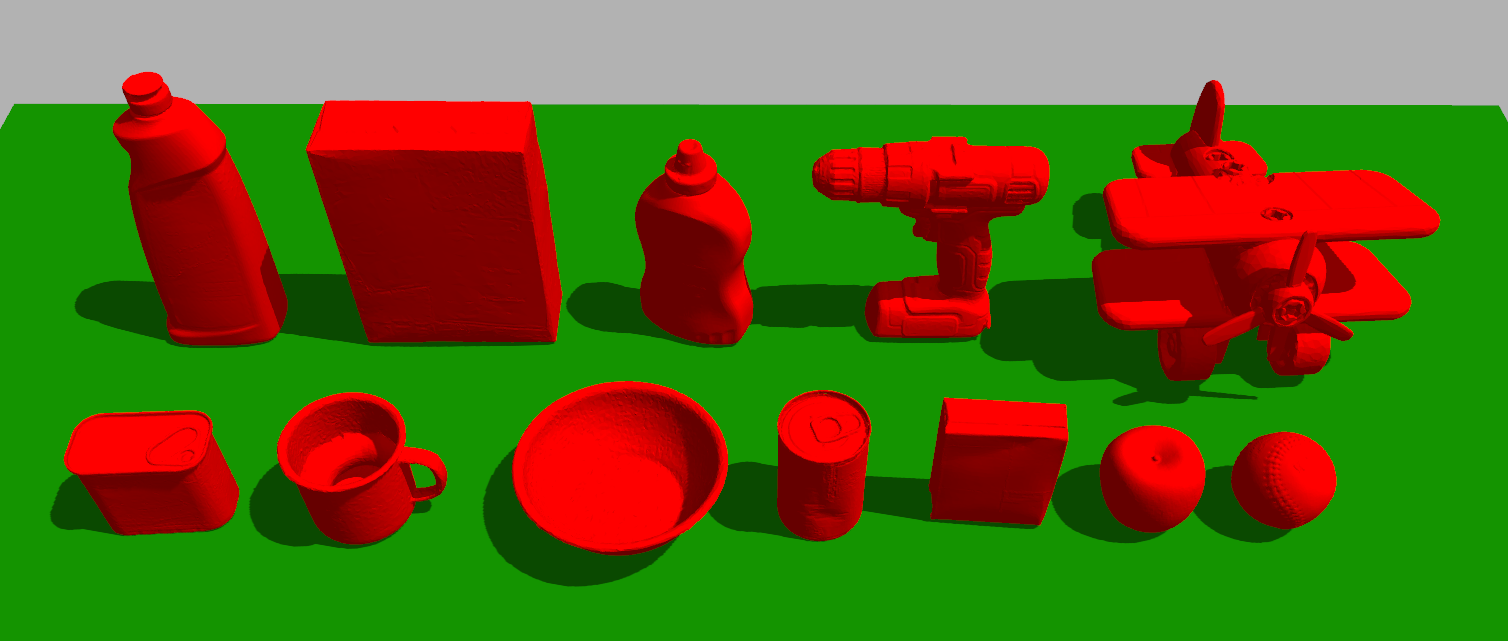}
    \caption{The set of object used for simulation testing. YCB object IDs : 3, 5, 7, 8, 10, 13, 21, 24, 25, 35, 55, 72-a}
    \label{fig_sim_exp_objects}
\end{figure}

The extensive testing in simulation was done on a set of 12 objects from the YCB dataset \citep{7254318} which are shown in Figure \ref{fig_sim_exp_objects}. To ensure consistency, we applied each algorithm to the exact same 100 poses for each object. This allowed us to produce a representative sample of a large number of points without biasing the dataset by using regular increments, while still giving each algorithm exactly identical conditions to work in. This was done by generating a set of 100 random values between 0 and 359 before testing began. To test a given policy with a given object, the object was spawned in Gazebo in a stable pose, with 0 degrees of rotation about the z-axis. The object was then rotated by the first of the random value about the z-axis, and the policy was used to search for a viable grasp. After the policy terminated, the object was reset, and rotated to the second random value, and so on.  

\begin{figure}[h]
    \centering
    \captionsetup{justification=centering}
    \includegraphics[width=\linewidth]{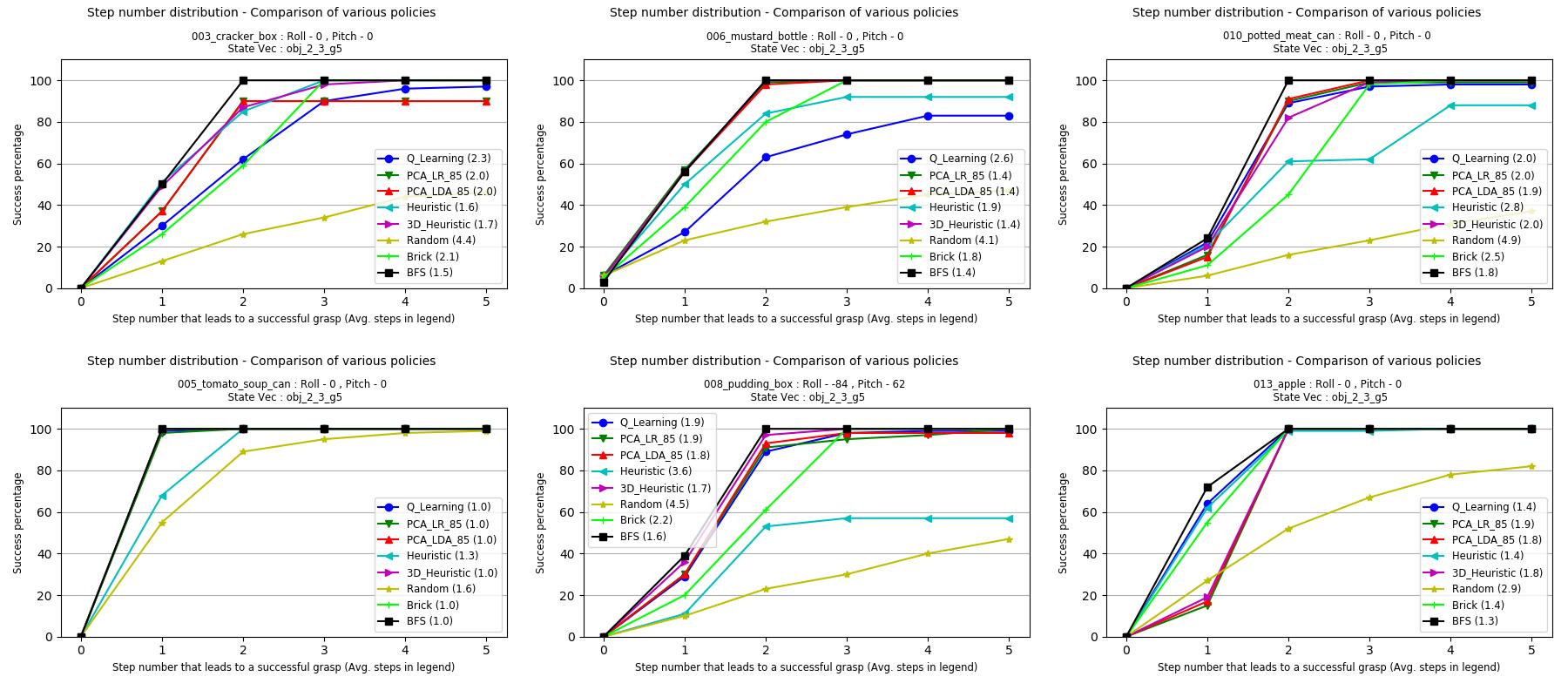}
    \includegraphics[width=\linewidth]{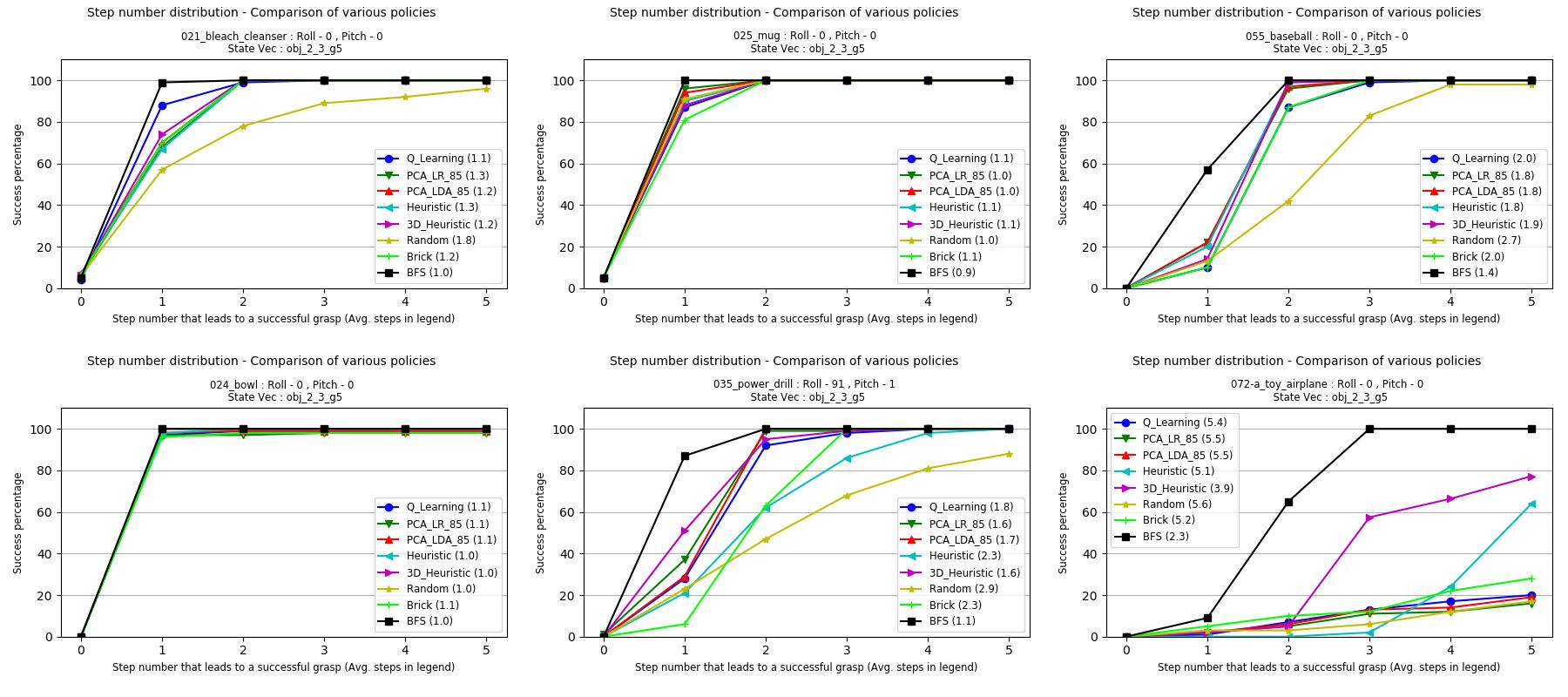}
    \caption{Simulation results for applying each approach to each object in 100 pre-set poses. Success is defined as reaching a view containing a grasp above a user defined threshold.  The number in parenthesis by the policy names in the legend is the average number of steps that policy took to find a grasp. For cases where no grasp was found, the step count was considered to be 6.}
    \label{fig_sim_res}
\end{figure}

The number of steps required to synthesise a grasp was recorded for each of the objects and its 100 poses tested. The success rate after each step for each object and the policies tested is shown in Figure \ref{fig_sim_res}. Each sub-image displays the fraction of poses a successful grasp has been reached for each policy on the same 100 pre-set poses for the given object. In object 025, for instance, the BFS found a working grasp on the first step for every starting pose, while all the other methods only found a grasp in the first step for a large majority of poses. By the second step every policy has found a working grasp for every tested pose of object 025.

\begin{figure}[h]
    \centering
    \captionsetup{justification=centering}
    \includegraphics[width=\linewidth]{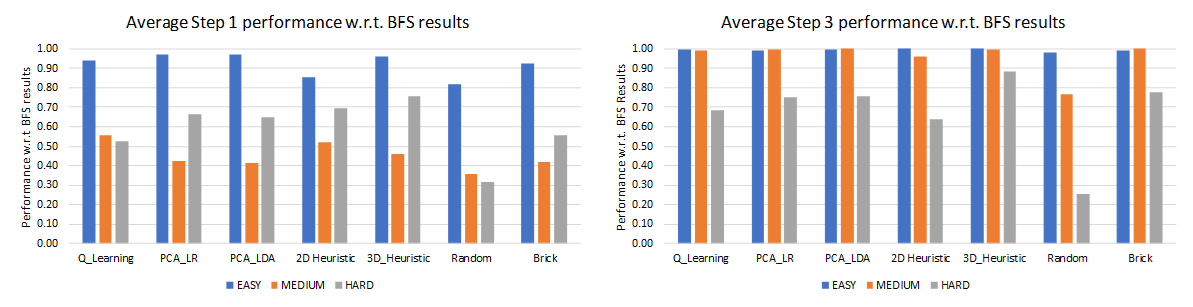}
    \caption{A comparison of performance of various policies for objects categorized into easy, medium and hard, for Step 1 and Step 3}
    \label{fig_sim_study_comp}
\end{figure}

The use of baseline policies i.e. random for the lower limit and BFS for the upper limit helped us in classifying the objects as easy, medium and hard in terms of how difficult is it to find a path that leads to a successful grasp. Objects are "Easy" when taking a step in almost any direction will lead to a successful grasp, and "Hard" when a low ratio of random to BFS searches succeed, suggesting very specific paths are needed find a grasp. Two objects with similar optimal and random performance will have similar numbers of paths leading to successful grasps, and so differences in performance between the two would be due to algorithmic failures, not inherent difficulty. The random to BFS ratio is used for the classification. For example, if the BFS result shows that out of 100 poses 40 poses have a successful grasp found in first step and a policy is only able to find a grasp at fist step for 10 poses, the policy is considered to have performed at 25\% of the optimal performance or in other words the ration would be 0.25. Objects with ratio at Step 2 $\leq$ 0.40 are considered hard, objects between 0.41 and 0.80 as medium, and objects with a ratio $>$ 0.80 as easy. With this criteria the test objects were classified as follows:
\begin{enumerate}
    \item Easy : Tomato soup can, Bowl, Mug
    \item Medium : Apple, Bleach cleanser, Power drill, Base ball
    \item Hard : Cracker box, Mustard Bottle, Pudding box, Potted meat can, Toy airplane
\end{enumerate}
With these object classifications, Figure \ref{fig_sim_study_comp} shows the performance of the policies for Step 1 and Step 3 using the policy to BFS ratio. 

Figures \ref{fig_sim_res} and \ref{fig_sim_study_comp} show that overall in simulation, the 3D Heuristic performed the best, followed by the self-supervised learning approaches, Q-Learning and the 2D Heuristic. For half of the objects we tested, the 3D Heuristic performed best, while for objects 003, 010, 013, 021, 025, and 055 another algorithm performed better. 

One reason the 3D Heuristic may be failing in some cases is that the heuristics are constrained to only considering the immediate next step. Our machine learning approaches can learn to make assumptions about several steps in the future, and so may be at an advantage on certain objects with complex paths. In addition, the optimistic estimations explained in Section~\ref{ssec:3d-heuristic-policy} would not always hold for all objects and cases. 
One reason for the machine learning techniques underperform for some cases may be due to the HAF representation, which creates a very coarse grained representation of the objects, obliterating fine details. A much finer grid size, or an alternative representation, could improve results.

We found that all methods consistently outperformed random, even on objects classified as hard. It is important to note that even brick policy was able to find successful grasps for all objects except for the toy airplane object (72-a), suggesting that incorporating active vision strategies even at a very basic level can improve the grasp synthesis for a object. 

The toy airplane object (72-a) deserves special attention as it was far and away the hardest object in our test set. It was the only object tested for which most algorithms did not achieve at least 80\% optimal performance by step 5, as well as having the lowest random to BFS ratio at step 5. 
We also saw (both here and in the real world experiments) that heuristic approaches performed the best on this extremely unusual object, while the machine learning based approaches all struggled to generalize to fit it.

Easy and Medium category objects come very close to optimal performance around step 3, as seen in Figure \ref{fig_sim_study_comp}. Given how small the possible gains on these simple objects can be, difficult objects should be the focus of future research. 
\FloatBarrier
\subsection{Comparison with the Information Gain Heuristic}
Using the same simulation setup the Information Gain Heuristic policy was compared to the 3D heuristic policy. The comparison results are shown in Table \ref{tbl:comparison}, where the number of viewpoints required was converted to the effective number of steps for 3D Heuristic for comparison. One step is the distance travelled to move to an adjacent viewpoint along the viewsphere in the discretized space with $v_{r}$ = 0.4m,  $v_{s}$ = 20\textdegree.

\begin{table}
    \centering
    \caption{Comparison between the exploration pattern employed by the Information Gain Heuristic and the 3D Heuristic's grasp weighted exploration.}
    \label{tbl:comparison}
    \includegraphics[width=0.9\linewidth]{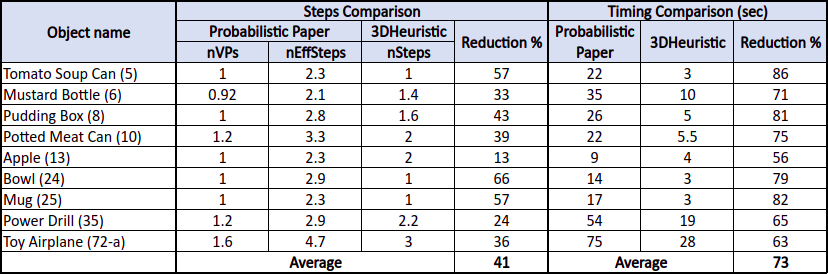}
\end{table}

We see an average of 41\% reduction in camera movement and with the 3D Heuristic policy, confirming our theory that only certain types of information warrants exploration and that by focusing on grasp containing regions we can achieve good grasps with much less exploration. As a side benefit, we also see a 73\% reduction in processing time with the 3D Heuristic policy, as it considers far fewer views in each step.

\FloatBarrier
\subsection{Real World Study}

\begin{figure}[h]
    \centering
    \captionsetup{justification=centering}
    \includegraphics[width=0.75\linewidth]{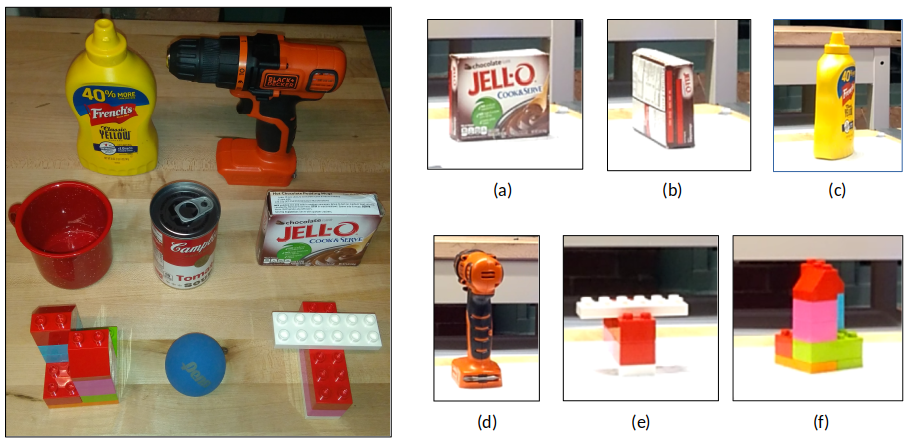}
    \caption{The left image shows the set of objects used for real world testing. On the right are the stable poses used for testing. (a) [YCB ID : 8] Stable Pose \#1, (b) [YCB ID : 8] Stable Pose \#2, (c) [YCB ID : 6] Stable Pose \#1, (d) [YCB ID : 35] Stable Pose \#1, (e) [Custom Lego 1] Stable Pose \#1, (f) [Custom Lego 2] Stable Pose \#1}
    \label{fig_lab_exp_objects}
\end{figure}

The real world testing was done on a subset of objects in simulation along with two custom objects built using lego pieces. The grasp benchmarking protocol in  \citep{Bekiroglu2020} was implemented to asses the grasp quality based of the five scoring parameters specified. The 3D Heuristic and the Q-Learning policies were selected and tested with the objects. The results for the tests performed are shown in Table \ref{tbl:lab_exp_res}. A total of 18 object-pose-policy combinations were tested with 3 trials for each and the average across the trails has been reported. The objects used along with their stable poses used for testing are shown in Figure \ref{fig_lab_exp_objects}. 

\begin{table}
    \centering
    \caption{A list of objects tested for 3DHeuristic and QLearning policies along with the benchmarking results}
    \label{tbl:lab_exp_res}
    \includegraphics[width=\linewidth]{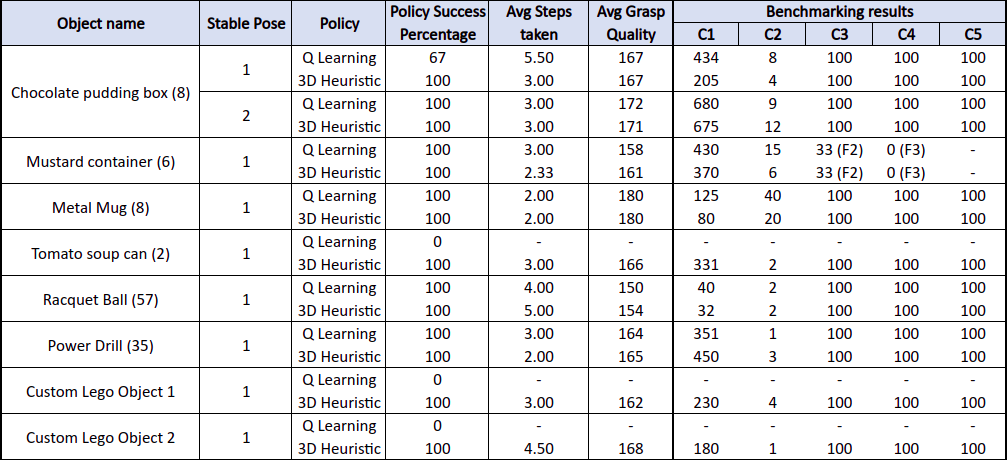}
\end{table}

\begin{figure}[h]
    \centering
    \captionsetup{justification=centering}
    \includegraphics[width=0.75\linewidth]{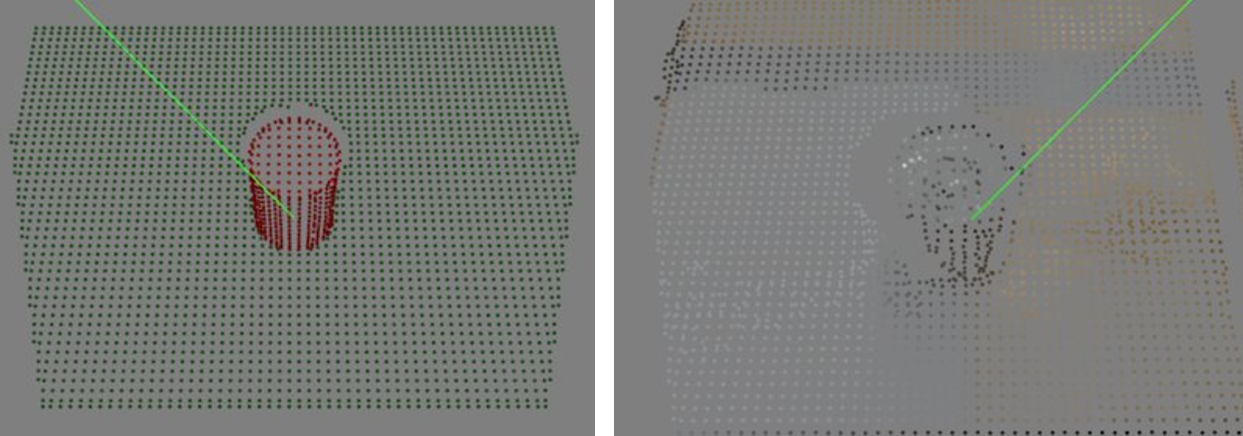}
    \caption{Difference between information captured by depth sensor in simulation (left) and real world (right)}
    \label{fig_sim_real_cam_diff}
\end{figure}

In real world trials, we found that the 3D heuristic works consistently, but the Q-Learning is at times unreliably. When run in simulation, the paths Q-Learning picks for the real-world objects produce successful grasps - the difference between our depth sensor in simulation and the depth sensor in the real world seems to be causing the disconnect. Figure \ref{fig_sim_real_cam_diff} shows the difference between the depth sensors in the two environments. The sensor in simulation is able to accurately see all the surfaces whereas in real world it fails to see the same amount of details. This also explains why more steps were required in the real world than in simulation. Nonetheless, the reliability of the 3D Heuristic demonstrates that simulated results can be representative of reality, although there are some differences.

\FloatBarrier

\section{Conclusions}
In this paper, we presented heuristic and data-driven policies to achieve viewpoint optimization to aid robotic grasping. In our simulation testing, we implemented a wide variety of active vision approaches and demonstrated that, for the YCB objects we tested, the 3D Heuristic outperformed both machine learning based approaches and naive algorithms. From our optimal search, we demonstrated that for most objects tested most approaches work well. We were able to identify that the most difficult object in our test set is not only dissimilar to our training objects, it is objectively more difficult to synthesize a grasp for. In the real world testing, we demonstrated that while sensor differences impacted all algorithms' performances, the heuristic based approach was sufficiently robust to generalize well to the real world while our machine learning based approaches were more sensitive to sensor noise. Finally, we demonstrated that prioritizing exploration of grasp-related locations produces both faster and more accurate policies. Future research should prioritize what we have identified as difficult objects over simple ones, as it is only in the more difficult objects that gains can be made and good policies discerned from poor ones.


\section*{Conflict of Interest Statement}

The authors declare that the research was conducted in the absence of any commercial or financial relationships that could be construed as a potential conflict of interest.

\section*{Author Contributions}
SN designed and implemented the 2D Heuristic. GB and SN designed the 3D Heuristic, and SN implemented it. SN developed the simulation with minor help from GB. GB ran the simulation testing. SN ran the real world testing. GB implemented the machine learning policies. SN and GB both analysed the data. BC provided research supervision and organized project funding. BC, SN, and GB collaborated on the writing.

\section*{Funding}
This work is partially supported by ``NRT-FW-HTF: Robotic Interfaces and Assistants for the Future of Work" with award number 1922761.




\bibliographystyle{frontiersinSCNS_ENG_HUMS} 
\bibliography{provisional}

\begin{thebibliography}{19}
\providecommand{\natexlab}[1]{#1}
\expandafter\ifx\csname urlstyle\endcsname\relax
  \providecommand{\doi}[1]{doi:\discretionary{}{}{}#1}\else
  \providecommand{\doi}{doi:\discretionary{}{}{}\begingroup
  \urlstyle{rm}\Url}\fi
\providecommand{\selectlanguage}[1]{\relax}
\providecommand{\bibAnnoteFile}[1]{%
  \IfFileExists{#1}{\begin{quotation}\noindent\textsc{Key:} #1\\
  \textsc{Annotation:}\ \input{#1}\end{quotation}}{}}
\providecommand{\bibAnnote}[2]{%
  \begin{quotation}\noindent\textsc{Key:} #1\\
  \textsc{Annotation:}\ #2\end{quotation}}

\bibitem[{Ammirato et~al.(2017)Ammirato, Poirson, Park, Kosecka, and
  Berg}]{Ammirato2017}
Ammirato, P., Poirson, P., Park, E., Kosecka, J., and Berg, A.~C. (2017).
\newblock A dataset for developing and benchmarking active vision.
\newblock In \emph{Proceedings - IEEE International Conference on Robotics and
  Automation}. 1378--1385.
\newblock \doi{10.1109/ICRA.2017.7989164}
\bibAnnoteFile{Ammirato2017}

\bibitem[{Arruda et~al.(2016)Arruda, Wyatt, and Kopicki}]{Arruda2016}
Arruda, E., Wyatt, J., and Kopicki, M. (2016).
\newblock Active vision for dexterous grasping of novel objects.
\newblock \emph{IEEE International Conference on Intelligent Robots and
  Systems} 2016-November, 2881--2888.
\newblock \doi{10.1109/IROS.2016.7759446}
\bibAnnoteFile{Arruda2016}

\bibitem[{{Bekiroglu} et~al.(2020){Bekiroglu}, {Marturi}, {Roa}, {Adjigble},
  {Pardi}, {Grimm} et~al.}]{Bekiroglu2020}
{Bekiroglu}, Y., {Marturi}, N., {Roa}, M.~A., {Adjigble}, K. J.~M., {Pardi},
  T., {Grimm}, C., et~al. (2020).
\newblock Benchmarking protocol for grasp planning algorithms.
\newblock \emph{IEEE Robotics and Automation Letters} 5, 315--322.
\newblock \doi{10.1109/LRA.2019.2956411}
\bibAnnoteFile{Bekiroglu2020}

\bibitem[{Caldera et~al.(2018)Caldera, Rassau, and Chai}]{Caldera2018}
Caldera, S., Rassau, A., and Chai, D. (2018).
\newblock Review of deep learning methods in robotic grasp detection.
\newblock \emph{Multimodal Technologies and Interaction} 2, 57.
\newblock \doi{10.3390/mti2030057}
\bibAnnoteFile{Caldera2018}

\bibitem[{Calli et~al.(2018{\natexlab{a}})Calli, Caarls, Wisse, and
  Jonker}]{Calli2018a}
Calli, B., Caarls, W., Wisse, M., and Jonker, P. (2018{\natexlab{a}}).
\newblock Viewpoint optimization for aiding grasp synthesis algorithms using
  reinforcement learning.
\newblock \emph{Advanced Robotics} 32, 1077--1089.
\newblock \doi{10.1080/01691864.2018.1520145}
\bibAnnoteFile{Calli2018a}

\bibitem[{Calli et~al.(2018{\natexlab{b}})Calli, Caarls, Wisse, Jonker, Wisse,
  and Jonker}]{Calli2018}
Calli, B., Caarls, W., Wisse, M., Jonker, P.~P., Wisse, M., and Jonker, P.~P.
  (2018{\natexlab{b}}).
\newblock Active vision via extremum seeking for robots in unstructured
  environments: Applications in object recognition and manipulation.
\newblock \emph{IEEE Transactions on Automation Science and Engineering} 15.
\newblock \doi{10.1109/TASE.2018.2807787}
\bibAnnoteFile{Calli2018}

\bibitem[{{Calli} et~al.(2015){Calli}, {Walsman}, {Singh}, {Srinivasa},
  {Abbeel}, and {Dollar}}]{7254318}
{Calli}, B., {Walsman}, A., {Singh}, A., {Srinivasa}, S., {Abbeel}, P., and
  {Dollar}, A.~M. (2015).
\newblock Benchmarking in manipulation research: Using the yale-cmu-berkeley
  object and model set.
\newblock \emph{IEEE Robotics Automation Magazine} 22, 36--52.
\newblock \doi{10.1109/MRA.2015.2448951}
\bibAnnoteFile{7254318}

\bibitem[{Calli et~al.(2011)Calli, Wisse, and Jonker}]{Calli2011}
Calli, B., Wisse, M., and Jonker, P. (2011).
\newblock Grasping of unknown objects via curvature maximization using active
  vision.
\newblock In \emph{IEEE International Conference on Intelligent Robots and
  Systems}. 995--1001.
\newblock \doi{10.1109/IROS.2011.6048739}
\bibAnnoteFile{Calli2011}

\bibitem[{Chollet et~al.(2015)}]{chollet2015keras}
[Dataset] Chollet, F. et~al. (2015).
\newblock Keras.
\newblock \url{https://keras.io}
\bibAnnoteFile{chollet2015keras}

\bibitem[{de~Croon et~al.(2009)de~Croon, Sprinkhuizen-Kuyper, and
  Postma}]{DeCroon2009}
de~Croon, G.~C., Sprinkhuizen-Kuyper, I.~G., and Postma, E.~O. (2009).
\newblock Comparing active vision models.
\newblock \emph{Image and Vision Computing} 27, 374--384.
\newblock \doi{10.1016/j.imavis.2008.06.004}
\bibAnnoteFile{DeCroon2009}

\bibitem[{Gallos and Ferrie(2019)}]{Gallos2019}
Gallos, D. and Ferrie, F. (2019).
\newblock Active vision in the era of convolutional neural networks.
\newblock In \emph{Proceedings - 2019 16th Conference on Computer and Robot
  Vision, CRV 2019}. 81--88.
\newblock \doi{10.1109/CRV.2019.00019}
\bibAnnoteFile{Gallos2019}

\bibitem[{Karasev et~al.(2012)Karasev, Chiuso, and Soatto}]{Karasev}
Karasev, V., Chiuso, A., and Soatto, S. (2012).
\newblock Controlled recognition bounds for visual learning and exploration.
\newblock In \emph{Advances in Neural Information Processing Systems}, eds.
  F.~Pereira, C.~J.~C. Burges, L.~Bottou, and K.~Q. Weinberger (Curran
  Associates, Inc.), vol.~25
\bibAnnoteFile{Karasev}

\bibitem[{Lakshminarayanan et~al.(2017)Lakshminarayanan, Pritzel, and
  Blundell}]{Lakshminarayanan2017}
Lakshminarayanan, B., Pritzel, A., and Blundell, C. (2017).
\newblock Simple and scalable predictive uncertainty estimation using deep
  ensembles.
\newblock \emph{Advances in Neural Information Processing Systems}
  2017-December, 6403--6414
\bibAnnoteFile{Lakshminarayanan2017}

\bibitem[{Paletta and Pinz(2000)}]{Paletta2000}
Paletta, L. and Pinz, A. (2000).
\newblock Active object recognition by view integration and reinforcement
  learning.
\newblock \emph{Robotics and Autonomous Systems} 31, 71--86.
\newblock \doi{10.1016/S0921-8890(99)00079-2}
\bibAnnoteFile{Paletta2000}

\bibitem[{Pedregosa et~al.(2011)Pedregosa, Varoquaux, Gramfort, Michel,
  Thirion, Grisel et~al.}]{scikit-learn}
Pedregosa, F., Varoquaux, G., Gramfort, A., Michel, V., Thirion, B., Grisel,
  O., et~al. (2011).
\newblock Scikit-learn: Machine learning in {P}ython.
\newblock \emph{Journal of Machine Learning Research} 12, 2825--2830
\bibAnnoteFile{scikit-learn}

\bibitem[{Rasolzadeh et~al.(2010)Rasolzadeh, Bj{\"{o}}rkman, Huebner, and
  Kragic}]{Rasolzadeh2010}
Rasolzadeh, B., Bj{\"{o}}rkman, M., Huebner, K., and Kragic, D. (2010).
\newblock An active vision system for detecting, fixating and manipulating
  objects in the real world.
\newblock \emph{International Journal of Robotics Research} 29, 133--154.
\newblock \doi{10.1177/0278364909346069}
\bibAnnoteFile{Rasolzadeh2010}

\bibitem[{Rusu and Cousins(2011)}]{Rusu_ICRA2011_PCL}
Rusu, R.~B. and Cousins, S. (2011).
\newblock {3D is here: Point Cloud Library (PCL)}.
\newblock In \emph{{IEEE International Conference on Robotics and Automation
  (ICRA)}} (Shanghai, China)
\bibAnnoteFile{Rusu_ICRA2011_PCL}

\bibitem[{Viereck et~al.(2017)Viereck, {Ten Pas}, Saenko, and
  Platt}]{Viereck2017}
Viereck, U., {Ten Pas}, A., Saenko, K., and Platt, R. (2017).
\newblock \emph{Learning a visuomotor controller for real world robotic
  grasping using simulated depth images}.
\newblock Tech. rep.
\bibAnnoteFile{Viereck2017}

\bibitem[{Wang et~al.(2020)Wang, Zhang, Zang, Liu, Ding, Yin et~al.}]{Wang2020}
Wang, C., Zhang, X., Zang, X., Liu, Y., Ding, G., Yin, W., et~al. (2020).
\newblock Feature sensing and robotic grasping of objects with uncertain
  information: A review 20, 1--28.
\newblock \doi{10.3390/s20133707}
\bibAnnoteFile{Wang2020}

\end{thebibliography}


\end{document}